\documentclass{article}

\usepackage{microtype}
\usepackage{graphicx}
\usepackage{subfigure}
\usepackage{booktabs} 
\usepackage{multirow}
\usepackage[numbers,sort&compress]{natbib}
\graphicspath{{./figs/}}

\usepackage{hyperref}

\newcommand{\tool}[1]{$\mathsf{#1}$}
\newcommand{\code}[1]{\texttt{\footnotesize#1}}

\usepackage[accepted]{icml2024}

\usepackage{amsmath}
\usepackage{amssymb}
\usepackage{mathtools}
\usepackage{amsthm}

\usepackage[capitalize,noabbrev]{cleveref}

\theoremstyle{plain}

\theoremstyle{definition}

\theoremstyle{remark}

\usepackage[textsize=tiny]{todonotes}

\begin{document}

\twocolumn[
\icmltitle{BetterV: Controlled Verilog Generation with Discriminative Guidance}



\icmlsetsymbol{equal}{*}

\begin{icmlauthorlist}
\icmlauthor{Zehua Pei}{cuhk}
\icmlauthor{Hui-Ling Zhen}{noah}
\icmlauthor{Mingxuan Yuan}{noah}
\icmlauthor{Yu Huang}{noah}
\icmlauthor{Bei Yu}{cuhk}
\end{icmlauthorlist}

\icmlaffiliation{cuhk}{The Chinese University of Hong Kong, Hong Kong SAR}
\icmlaffiliation{noah}{Noah’s Ark Lab, Huawei, Hong Kong SAR}

\icmlcorrespondingauthor{Bei Yu}{byu@cse.cuhk.edu.hk}

\icmlkeywords{Machine Learning, ICML}

\vskip 0.3in
]



\printAffiliationsAndNotice{}  

\begin{abstract}

Due to the growing complexity of modern Integrated Circuits (ICs), there is a need for automated circuit design methods.
Recent years have seen rising research in hardware design language generation to facilitate the design process.
In this work, we propose a Verilog generation framework, BetterV, which fine-tunes the large language models (LLMs) on processed domain-specific datasets and incorporates generative discriminators for guidance on particular design demands.
The Verilog modules are collected, filtered and processed from internet to form a clean and abundant dataset.
Instruct-tuning methods are specially designed to fine-tune the LLMs to understand the knowledge about Verilog.
Furthermore, data are augmented to enrich the training set and also used to train a generative discriminator on particular downstream task, which leads a guidance for the LLMs to optimize the Verilog implementation.
BetterV has the ability to generate syntactically and functionally correct Verilog, which can outperform GPT-4 on the VerilogEval benchmark.
With the help of task-specific generative discriminator, BetterV can achieve remarkable improvement on various electronic design automation (EDA) downstream tasks, including the netlist node reduction for synthesis and verification runtime reduction with Boolean Satisfiability (SAT) solving.

\end{abstract}
\section{Introduction}

Large language models (LLMs) are leading the world because of their strong capability on generating and understanding natural language at a massive scale, which makes their potential applications and benefits for various domains and tasks.
On the field of coding, LLMs can act as a seasoned assistant for developers to give advise on programming, find and fix bugs on the code and even generate the whole code with descriptions~\cite{chen2021evaluating,nijkamp2023codegen2}.

Electronic design automation (EDA) is a set of softwares and services for designing integrated circuits (ICs), which work together in design flow to conceive and analyze the circuit designs.
The slowing down of the Moore's law puts an increasing pressure on EDA, and there is an emergent need to further improve and automate the design flow.
Therefore, it is expected to incorporate LLMs into EDA flow.
LLMs on EDA have already achieved remarkable success on various tasks~\cite{liu2023chipnemo,he2023chateda}.

Hardware design languages (HDLs), such as Verilog and VHDL, describe the hardware design at the very beginning of the design flow, which play an important role in the EDA flow and have strong influence on the following stages.
However, writing the HDL is time-consuming and bug-prone, which makes it more expensive for today's complex ICs.
Therefore, it is promising to utilize the LLMs to automatically generate the desired HDL.
Recently, there have been several works focus on Verilog generation~\cite{thakur2023benchmarking, dehaerne2023deep}.
However, these works only focus on refining LLMs with selected datasets, neglecting the direct integration of syntactic or functional correctness in the model's development. 
Moreover, they insufficiently address pivotal downstream tasks in the EDA process, which should be crucial evaluations for the generated Verilog.

Despite that we expect that LLMs can play an important role in Verilog generation, there still are some challenges that we need to figure out.
Firstly, the complex and strict requirements of hardware designs restrain the LLMs to learn and understand the knowledge related to the Verilog.
Secondly, there are limited Verilog resources available in the world, which always leads to the problems of overfitting and data bias for LLMs fine-tuning.
Moreover, taking the complicated and various downstream tasks into consideration further makes it a challenge, since in the industrial production the optimize of downstream targets from Verilog design to physical implementation is various and can't be accomplished in an action, which always need many loops in the industrial production. 
Therefore, it's difficult for LLMs to look ahead and it's also not practical to fine-tune the LLMs for each downstream task.

The controllable text generation contains techniques to train extra discriminators to guide the LLMs on desired directions~\cite{ scialom2020discriminative}.
However, these works only pay attention to control the natural language text representation in their development.
In this work, we try tapping into the potential to use this technique on optimization tasks, which is more difficult and different from simply treating the text representation.

In this work, we propose a framework, BetterV, for Verilog generation, by instruct-tuning the LLMs on our processed datasets and incorporating generative discriminators to optimize the Verilog implementation on various downstream tasks.
We utilize the alignment between Verilog and the C program to help the LLMs understand the Verilog effectively.
At the same time, data augmentation is proposed to solve the Verilog data scarce issue.
What's more, we consider that there are many implementations of Verilog that always have different performance on PPA (Power, Performance, Area) or verification runtime.
Therefore, the problems in downstream tasks are transformed to how to optimize the Verilog implementation.
The generative discriminator is then employed to guide the LLMs to generate or modify Verilog implementation directly from natural language and be as friendly to do downstream tasks as possible at the same time, thereby effectively reducing the number of iterations on industrial production.
With the above techniques, BetterV successfully teaches the LLMs to understand domain-specific knowledge and can adapt to any Verilog related downstream tasks.

The contributions of this paper are summarized as follows:
\begin{itemize}
    \item BetterV represents a groundbreaking development as the first endeavor to apply controllable text generation to engineering optimization challenges, specifically in optimizing downstream tasks in Electronic Design Automation (EDA). 
    This approach not only introduces an innovative and promising research trajectory in EDA but also holds potential for application in optimization issues across various other domains.
    \item 
    BetterV marks a pioneering advancement as the first downstream task-driven method for Verilog generation. 
    Our experiments employing various generative discriminators on specific Electronic Design Automation (EDA) tasks have demonstrated notable effectiveness. 
    This innovative approach is characterized by its task-specific discriminator guidance, enhancing both its training efficiency and practical utility.
    \item Utilizing fine-tuned 6.7B/7B-parameter LLMs, without the application of prompt-engineering strategies, BetterV demonstrates the capacity to generate syntactically and functionally correct Verilog, which surpasses GPT-4 when evaluated on the VerilogEval benchmark.
    \item BetterV offers a versatile solution for data augmentation, tailored to meet diverse specifications in Verilog implementation. 
    This approach addresses the issue posed by the scarcity of Verilog resources effectively.
\end{itemize}

\section{Related Works}

In this section, we briefly introduce some advancements and applications of LLMs for Verilog generation in \Cref{sec:relate_1}.
We also discuss the development of discriminator-guided controllable generation in \Cref{sec:relate_2}.

\subsection{LLMs for Verilog Generation}
\label{sec:relate_1}

Large language models (LLMs) have shown remarkable performance on code generation with either training a model from the beginning or fine-tuning a pre-trained model~\cite{nijkamp2023codegen2,roziere2023code}.
The success of LLMs on code generation arouses the interest of study of LLMs on hardware design.
As a widely recognized hardware description language (HDL), the generation of Verilog using LLMs has garnered significant attention and undergone extensive exploration.
Some studies pay attention to construct customized datasets for the fine-tuning of LLMs, such as \cite{thakur2023benchmarking} and \cite{dehaerne2023deep}, who collect Verilog from the internet and process the data prior to training.
VerilogEval~\cite{liu2023verilogeval} and RTLCoder~\cite{liu2023rtlcoder} further study the importance of problem descriptions and then generate various problem-answer pairs as dataset.
RTLCoder also considers the quality feedback on different data by ranking them with scores during the training.
Researchers also try utilizing the prompt-engineering techniques to enhance the generation ability, such as the self-planning used in RTLLM~\cite{lu2023rtllm}.
VerilogEval~\cite{liu2023verilogeval} and RTLLM~\cite{lu2023rtllm} build benchmarks to evaluate the generated Verilog on their functional or syntactic correctness.

\subsection{Discriminator-guided Controllable Generation}
\label{sec:relate_2}

Controlling LLMs has been widely explored during recent years.
Class-conditional language models (CCLMs), such as CTRL~\cite{keskar2019ctrl}, aim at controlling the generation by appending a control code in the beginning of training sequences.
Discriminator-guided controllable generation is an important technique used for controllable generation, which combines a discriminator with the generative LLMs.
The constraints are modeled by calculating the conditional probability on each class and related to the next-token probabilities by Bayes rule.
\cite{holtzman2018learning} and \cite{scialom2020discriminative} predict the labels by feeding each candidate next token into a discriminator and hence guiding the generation to desired directions.
PPLM~\cite{Dathathri2020Plug} further employs a forward and backward process with the gradient from a discriminator to update the latent states of LLMs to guide the target generation.
In order to reduce the cost of using discriminator for each possible next tokens, GeDi~\cite{krause2020gedi} contrastively trains the CCLMs as generative discriminators to guide the generation during the decoding.
DEXPERTS~\cite{liu2021dexperts} attempts to further improve the performance by incorporating expert and anti-expert during the decoding.

\section{Algorithm}

In this section, we describe the framework of BetterV from an overview to the details of the instruct-tuning and the implementation of generative discriminator.

\begin{figure}[ht]
    \centering
    \includegraphics[width=0.99\linewidth]{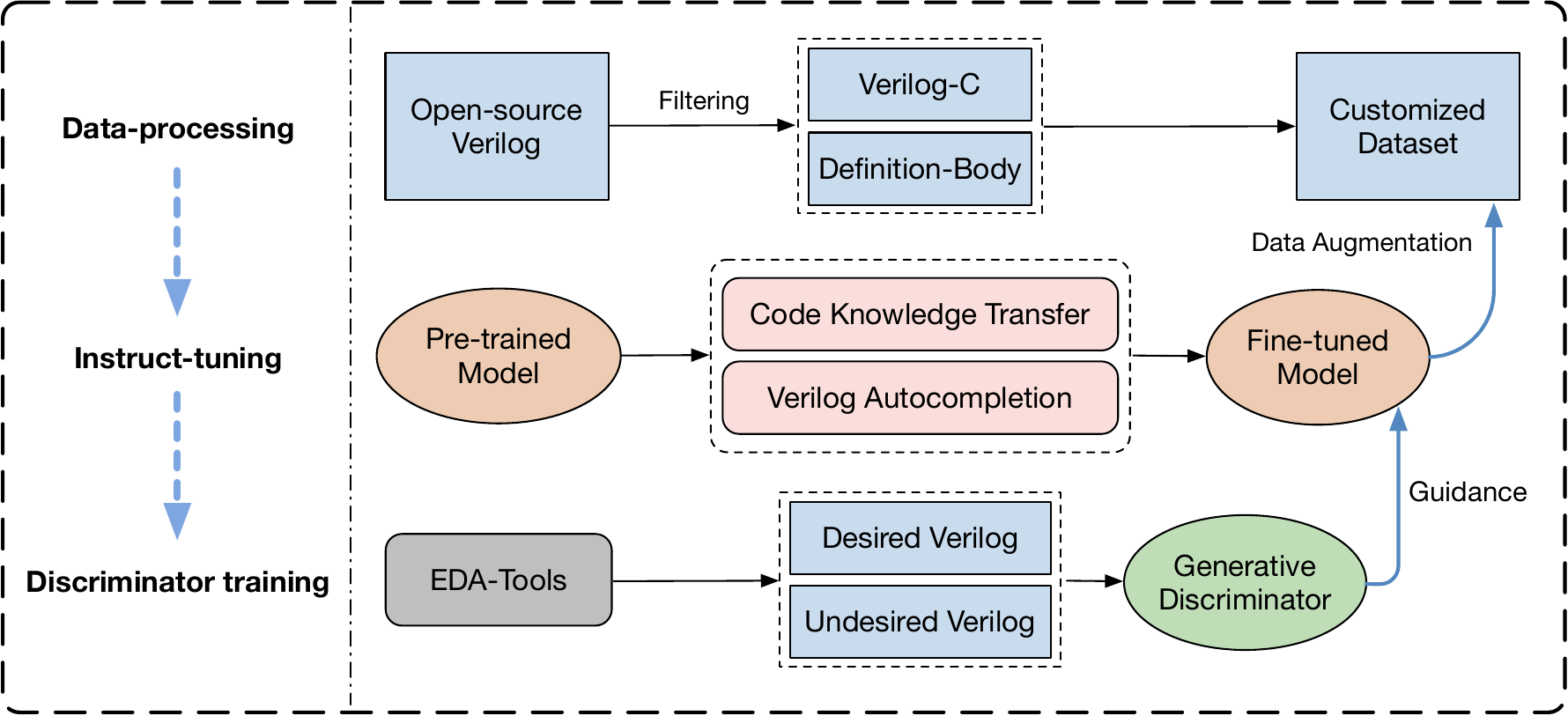} 
    \caption{
    The overview of BetterV.
    }
    \label{fig:overview}
\end{figure}

\subsection{Framework Overview}

BetterV conducts instruct-tuning for domain-specific understanding and combines generative discriminators for guidance on different EDA downstream tasks.
The framework overview is demonstrated in \Cref{fig:overview}.
Our approach first constructs customized dataset from open-source Verilog in \Cref{sec:data} before the instruct-tuning stage.
During instruct-tuning, various instruction problems are employed to teach the LLMs domain-specific knowledge as in \Cref{sec:finetune}.
After that, we implement data augmentation to further enrich the dataset and prepare the labels for training the generative discriminator in \Cref{sec:aug}.
Finally in \Cref{sec:dis}, the generative discriminator is trained followed by a hybrid generative-discriminative loss and then guides the LLMs through Bayes rule.

In the rest of this section, we present the details of BetterV in guiding the LLMs on downstream EDA tasks optimization.

\subsection{Instruct-Tuning Data-Processing}
\label{sec:data}

Inspired by~\cite{dehaerne2023deep}, we collect open-source repositories in GitHub that contain Verilog or SystemVerilog. 
At the same time, the repository licenses are also checked to permit modification and distribution.
We also filter out the auto-generated files that are mostly repetitive and the non-permissive files that contradict the licenses.
As in~\cite{dehaerne2023deep}, the files with too much or less lines are filtered.
These files are then processed by extracting the Verilog modules and functions with regular expressions.
We further analyze the extracted contents by measuring the token number after encoding them with the tokenizer and then remove the contents exceeding a pre-defined maximum token number.
The reason is because the contents that exceed the maximum token number will be truncated by the tokenizer during training and hence it cannot provide enough meaningful information with incomplete contents.
Finally, we filtered out the Verilog modules that have syntactic error.

In order to further support our instruct-tuning tasks, we implement two kinds of processing on the collected contents.
The first one is that we use a V2C tool to translate Verilog into C~\cite{mtk2016}, then the hardware design information represented in Verilog is also transferred into the C program.
The dataset is then appended by the translated C programs, which are able to be translated from the Verilog and also don't exceed the maximum token number.
Secondly, we split the Verilog modules and functions into their definition (with the module header, input and output definition) and body.

Finally, our constructed dataset contains a set of Verilog-C pairs and a set of Verilog definition-body pairs.

\subsection{Domain-specific Instruct-Tuning}
\label{sec:finetune}

The fine-tuning of LLMs is essentially important since it basically decides how well the domain-specific knowledge is learnt by the LLMs and how familiar the LLMs is with the customized tasks.
It is relatively more important for the task of Verilog generation, the reason is that the existed corpus of Verilog is much less than general code programs such as C and Python.
The lack of corpus of Verilog not only indicates that the pre-trained LLMs learn less knowledge about it, but also upgrades the importance to transfer more domain-specific knowledge through fine-tuning.
Previous works implement the fine-tuning by generating the problem description with LLMs first, and then using the LLMs to further generate the corresponding answer, i.e. the Verilog module. 
However, such machine-generated description is always verbose and fallible, which fails to transfer useful and reliable knowledge during fine-tuning.
Moreover, since such generation is done by LLMs inference, it will results in time-consuming and occupancy of resources, which largely increases the implementation cost.

\begin{figure}[tb!]
    \centering
    \includegraphics[width=0.97\linewidth]{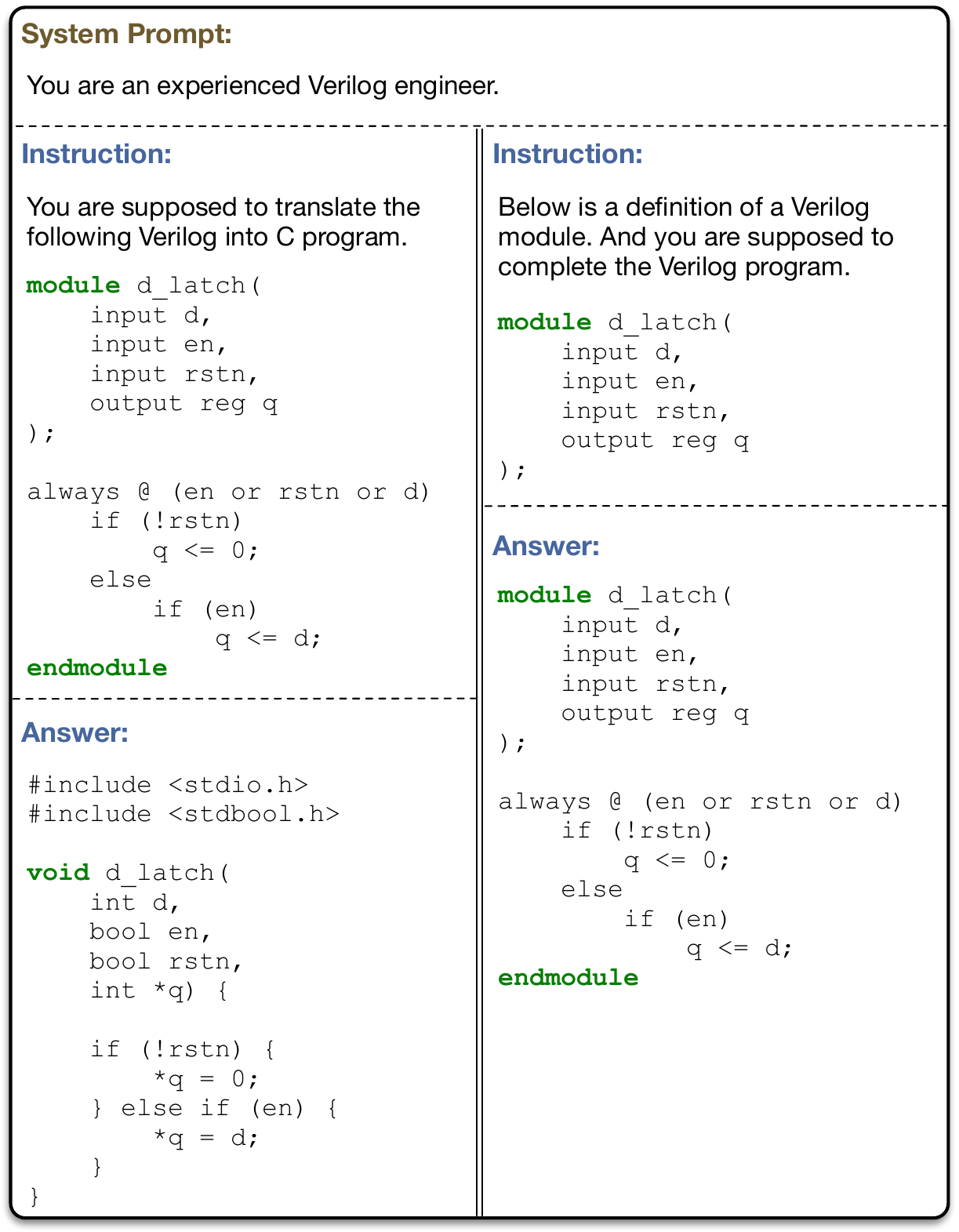} 
    \caption{
    The examples for instruct-tuning. We take a simple \code{d\_latch} Verilog module as example. The left and right part indicate the code knowledge transfer and Verilog autocompletion, respectively.
    }
    \label{fig:instruction}
\end{figure}

To solve this issue, we propose domain-specific instruct-tuning, where examples are shown in \Cref{fig:instruction}.
Firstly, we introduce a novel alignment method that maps Verilog code to C, facilitating LLM's understanding of Verilog.
We make use of the abundant knowledge already learnt by LLMs on general code program, i.e. the C program.
Then we construct an instruction that asks the LLMs to translate the Verilog into C and also translate from C to Verilog, by which LLMs can learn to understand Verilog from the implementation of C.
With the help of V2C tool, the dataset is easy to be obtained as described in last section.
Besides this, in order to improve the performance of LLMs to follow the Verilog generation instruction, the instruction on Verilog autocompletion is constructed, where the module definition is in the instruction and the completed module is the answer.
It should be noted that we don't add extra natural language description in instruct tuning since the implementation of C program already acts as functional description to teach the LLMs.

To be noticed, the LLMs follow the an auto-regressive manner, where a LLM with parameter $\theta$ predicts the probability of a sequence $x = \{x_1, ..., x_n\}$ with both instruction and answer and factorizes it using the chain rule of probability as follows~\cite{keskar2019ctrl}:
\begin{equation}
\label{eq:1}
    p_{\theta}(x) = \prod_{t = 1}^{n} p_{\theta}(x_t|x_{<t}).
\end{equation}
The generation of LLMs iteratively samples from $p_{\theta}(x_t|x_{<t})$ and then takes $x_t$ back into the input for next prediction.
Therefore, the LLMs are trained to minimize the negative log-likelihood on a set of training sequences D $\{x^1, ..., x^{|D|} \}$:
\begin{equation}
    \mathcal{L} = -\frac{1}{|D|}\sum_{i=1}^{|D|}\frac{1}{n}\sum_{t = 1}^{n}\log p_{\theta}(x^i_t|x^i_{<t}).
\end{equation}

\subsection{Data Augmentation}
\label{sec:aug}

Verilog is scarce compared with other general code programs, which makes the LLMs easy to overfitting and hence decrease the performance.
Therefore, in this section we consider to use the LLMs to create synthetic Verilog to increase the diversity and size of the training set.
After instruct-tuning the LLM, it has the ability to generate syntactically correct Verilog.
Therefore, it is convenient to directly use the fine-tuned LLMs to do data augmentation.
We first make use of the module head that we collected in the data processing stage.
Then we construct instructions to ask the LLMs to directly complete the module.
With high temperature set during generation, our LLMs can produce diverse augmented Verilog modules and forms a preliminary synthetic dataset.
With the generated Verilog modules, we finally employ the EDA tool to check the syntactic correctness and filter the syntax error modules.
By doing the data augmentation, the robustness and generalization of the fine-tuned model are improved.
Moreover, since all the remaining Verilog modules are syntactic correctness, appending the synthetic dataset can further enhance the ability to generate syntactically correct Verilog.

Besides augmenting data for instruct-tuning, we need to prepare data for training the discriminator for downstream tasks.
Although we can already classify the Verilog modules in our dataset by specific EDA downstream tools, we still need more data.
In order to make the discriminator better distinguishing the difference between our desired and undesired Verilog, we further utilize our LLMs to generate Verilog modules by completing the module heads.
Then these augmented modules are also labeled by the EDA tools, which is determined by special syntax or hardware attributes.
Sometimes the attributes that we consider for Verilog implementation have absolutely true or false, e.g. the syntactically and functional correctness, then in this case the desired or undesired are directly labeled.
Sometimes they only has relative good or bad, e.g. whether the netlist nodes are lower after synthesis, and in this case the desired or undesired label according to specific target compared with the references.
Hence for this case we always keep a reference Verilog and then generate the corresponding data.

\subsection{Generative Discriminator}
\label{sec:dis}


After preparing the training data, we discuss how to train a generative discriminator and then use it to guide the LLMs to generate better Verilog.
As classical class-conditional language models (CC-LMs), the LLMs are conditioned on an attribute variable, which is expressed as a control code $c$ that assigned at the beginning of the input sequences.
Therefore, the probability that the LLMs to predict becomes $p_{\theta}(x|c)$, which is the conditional probability of $x$ on the attribute $c$ and its factorization is similar to \Cref{eq:1}:
\begin{equation}
    p_{\theta}(x|c) = \prod_{t = 1}^{n} p_{\theta}(x_t|x_{<t},c).
\end{equation}
For generative discriminator, we have binary cases to represent opposite attributes, i.e. a control code $c$ and an anti-control code $\Bar{c}$~\cite{krause2020gedi}, and the inputs are respectively conditioned by them as $p_{\theta}(x|c)$ and $p_{\theta}(x|\Bar{c})$ to guide the LLMs on $p_{LLM}(x)$.
In our domain-specific scenarios, the preceding codes indicate which kind of Verilog attribute we desired or undesired.
As described in \Cref{sec:aug}, we expect the generative discriminator can generate correct and better Verilog implementation, and their labels are corresponding to the conditioned attributes.

\begin{figure}[tb!]
    \centering
    \includegraphics[width=0.9\linewidth]{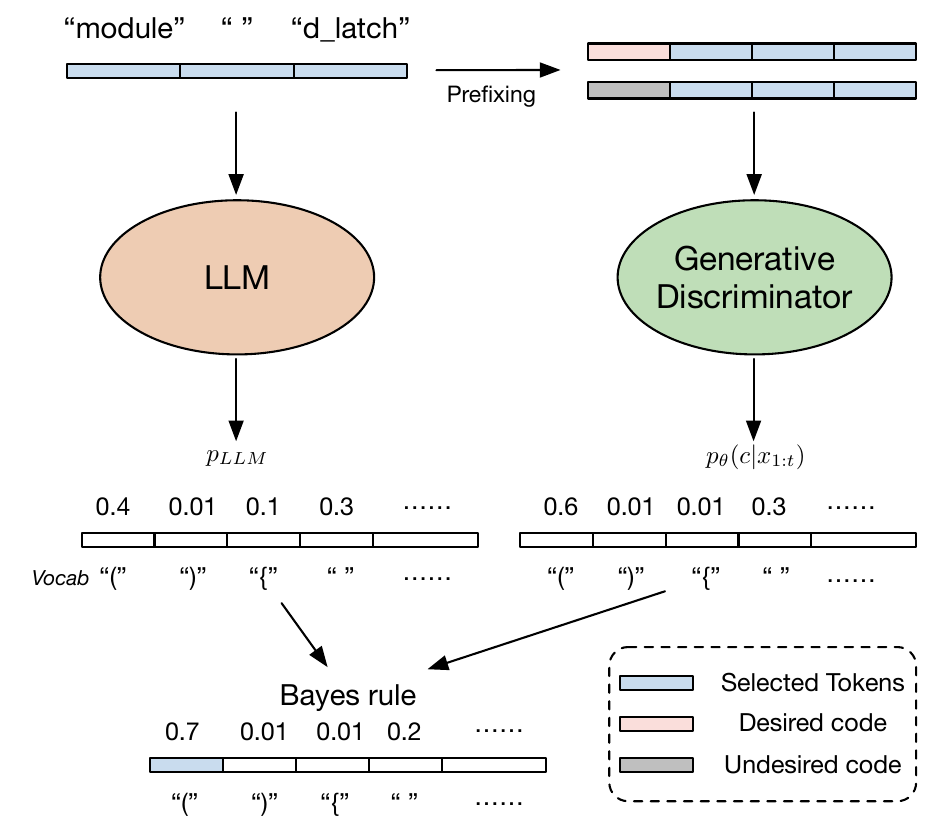} 
    \caption{
    An example shows the guidance from generative discriminator.
    }
    \label{fig:guidance}
\end{figure}

The discriminator is first trained to predict the next token for each attribute during generation on a set of training sequences D $\{x^1, ..., x^{|D|} \}$ with the generative loss $L_g$:
\begin{equation}
    \mathcal{L}_g = -\frac{1}{|D|}\sum_{i=1}^{|D|}\frac{1}{n}\sum_{t = 1}^{n}\log p_{\theta}(x^i_t|x^i_{<t}, c'),
\end{equation}
where $c' \in \{ c,\Bar{c} \}$.
With the predicted $p_{\theta}(x_{1:t}|c)$ and $p_{\theta}(x_{1:t}|\Bar{c})$, Bayes rule is used to compute the probability that each next token $x_t$ belongs to the labeled class:
\begin{equation}
    \begin{aligned}
        p_{\theta}(c_y|x_{1:t}) &= \frac{p(c_y)(\prod_{j=1}^{t}p_{\theta}(x_j|x_{<j},c_y))^{\alpha/t}}{\sum_{c'\in \{ c,\Bar{c} \}} p(c')\prod_{j=1}^{t}(p_{\theta}(x_j|x_{<j},c'))^{\alpha/t}}  \\
        &= \frac{p(c_y)p_{\theta}(x_{1:t}|c_y)^{\alpha/t}}{\sum_{c'\in \{ c,\Bar{c} \} } p(c')p_\theta(x_{1:t}|c')^{\alpha/t} },
    \end{aligned}
\end{equation}
where $c_y\in \{ c,\Bar{c} \}$ is the label for the sequence $x$, $\alpha$ is a learnable scale parameter, $p(c) = \frac{e^{b_c}}{\sum_{c'} e^{b_{c'}}}$ with $b_c$ is also a learnable bias for control code $c$ and the probabilities are normalized by the current sequence length $t$.
Then a key point is raised that the discriminator need to be trained that can distinguish the class of each sequence, which means to discriminatively train the class-conditional generative models.
Therefore, given the data with both the sequences D $\{x^1, ..., x^{|D|} \}$ and the corresponding labels $\{c_y^1, ..., c_y^{|D|} \}$, the discriminative loss $\mathcal{L}_d$ is defined as follows:
\begin{equation}
    \mathcal{L}_d = -\frac{1}{|D|} \sum_{i=1}^{|D|} \log p_{\theta}(c_y^i|x_{1:n}^i).
\end{equation}
Finally, the total loss function of generative discriminator training is defined as follows:
\begin{equation}
    \mathcal{L}_{total} = \lambda \mathcal{L}_g + (1- \lambda) \mathcal{L}_d,
\end{equation}
where $\lambda$ is a hyper-parameter to balance the weight between generative loss and discriminative loss.

After training the generative discriminator, we can use it to guide the sampling of LLMs.
In \Cref{fig:guidance}, we simply show how this works with an example.
The weighted decoding with Bayes rule is employed to guide the generation:
\begin{equation}
    p_w(x_t|x_{<t},c) \propto p_{LLM}(x_t|x_{<t})\ p_{\theta}(c|x_t,x_{<t})^w,
\end{equation}
where $w$ is a hyper-parameter to control the influence of the weighted conditional probability.
With the weighted decoding, all the potential next tokens $x_t$ in the vocabulary are updated.
Moreover, some filtering methods on the tokens before sampling are utilized.
The insight is to maintain the high probability tokens and filter out the low probability tokens.
First, we define the complete vocabulary set as $\mathcal{V}$.
Then we rank all the tokens $x_t$ on the probability $p_{\theta}(c|x_t,x_{<t})$ to form a new set $\mathcal{V}_{rank}$, and we maintain the top tokens that have the minimum number $m$ such that:
\begin{equation}
    \sum_{x_t\in \mathcal{V}_{rank}[1:m]} p_w(x_t|x_{<t},c) \geq \rho,
\end{equation}
which is a minimum of at least $\rho$ in cumulative probability mass on $p_w(x_t|x_{<t},c)$ and the maintained set is defined as $\mathcal{V}_m = \mathcal{V}_{rank}[1:m]$.
Furthermore, to avoid the case that the high $p_{\theta}(c|x_t, x_{<t})$ are filtered out, the tokens that $p_{\theta}(c|x_t,x_{<t}) > \tau$ are maintained and form the set $\mathcal{V}_{\tau}$.
Therefore, the tokens that we keep before the sampling of generation are given by $\mathcal{V}_k = \mathcal{V}_{\tau} \cup \mathcal{V}_{m}$.

With this generative discriminator training pipeline and the equipped downstream task-specific augmented data, we can model any optimization scenario related to Verilog implementation and train a distinct discriminator to guide towards our desired directions.
At the same time, the loops in the industrial production and the manual cost can be reduced effectively by jointly considering the corresponding downstream constraints at the phase of Verilog generation.

\section{Experiments}

In this section, we first introduce the experimental setting as well as the evaluation metrics in \Cref{sec:exp_1}.
Then in \Cref{sec:exp_2} we present our results on functional correctness and compared with other methods, followed by further results with different downstream tasks incorporated with the discriminator (including Synthesis Nodes Reduction in \Cref{sec:exp_3} Verification Runtime Reduction in \Cref{sec:exp_4}) and ablation study in \Cref{sec:exp_5}.

\subsection{Experimental Setting}
\label{sec:exp_1}

We fine-tune the CodeLlama-7B-Instruct~\cite{roziere2023code} and DeepSeek-Coder-6.7b-Instruct~\cite{guo2024deepseek} as generative LLMs, and fine-tune TinyLlama~\cite{zhang2024tinyllama} and DeepSeek-Coder-1.3b-Instruct~\cite{guo2024deepseek} as generative discriminator, respectively.
For CodeQwen1.5-7B-Chat~\cite{bai2023qwen}, we can only fine-tune it since there is no suitable smaller size version of it.
BetterV is trained with the help of DeepSpeed ZeRO~\cite{rajbhandari2020zero} and LoRA~\cite{hu2021lora} on the DeepSpeed-Chat~\cite{yao2023dschat}.
The experiments are conducted on a machine with two NVIDIA Tesla V100S PCIe 32 GB graphics cards with CUDA driver 11.4.

We employ the VerilogEval~\cite{liu2023verilogeval}, which comprises various problems with either machine-generated or human-crafted, as our evaluation benchmark.
Following VerilogEval, we measure the Verilog functional correctness with simulation through the pass@k metric with unbiased estimator:
\begin{equation}
    \text{pass@}k := \mathop{\mathbb{E}}_{\text{problems}} \left[\frac{1 - \binom{n-c}{k}}{\binom{n}{k}}\right],
\end{equation}
where $n \leq k$ samples are generated per problem and a problem is solved if any of the k samples passes the unit tests.

Due to the Plug-and-Play nature of the discriminator, BetterV can work on various downstream tasks.
For other downstream tasks besides the functional correctness, we can still utilize the pass@k metric and the problems in VerilogEval but the correctness is measured by different tools in \tool{Yosys}~\cite{Yosys} according to different downstream task.
In our experiments, we sample n = 20 code completions per problem for each downstream task and measuring pass@k with k = 1,5,10.

For all the models, we employ the Adam optimizer~\cite{kingma2014adam} with $\beta_1$ = 0.9 and $\beta_2$ = 0.95 and the cosine learning rate decay~\cite{loshchilov2016sgdr} to schedule our learning rate.
For the generative LLMs fine-tuning process, we train it for 4 epochs using an initial learning rate of 9.65e-6 with a batch size of 4.
For the generative discriminator, we train it for 3 epochs using an initial learning rate of 9.65e-6 with a batch size of 8.
The LoRA dimension for both LLMs and discriminator is set as 128.
For different downstream tasks they have different sensitivity for the incorporation of generative discriminators, hence the setting of $\lambda$, $w$, $\rho$ and $\tau$ are different and task-sensitive.
We initialize the value of learnable scale $\alpha$ as $1$ and bias $b$ as $0$.

\begin{table}[tb]
    \centering
    \caption{Comparison of functional correctness on VerilogEval.}
    \label{tab:func_1}
    \renewcommand{\arraystretch}{1.18}
    \resizebox{1.0\linewidth}{!}{
    \begin{tabular}{c|ccc|ccc}
        \toprule
        \multirow{2}{*}{Model} & \multicolumn{3}{c|}{VerilogEval-machine}    & \multicolumn{3}{c}{VerilogEval-human} \\
         & pass@1 & pass@5 & pass@10 & pass@1 & pass@5 & pass@10 \\
         \midrule
        GPT-3.5 & 46.7 & 69.1 & 74.1 & 26.7 & 45.8 & 51.7 \\
        GPT-4 & 60.0 & 70.6 & 73.5 & 43.5 & \textbf{55.8} & \textbf{58.9} \\
        CodeLlama & 43.1 & 47.1 & 47.7 & 18.2 & 22.7 & 24.3 \\
        DeepSeek & 52.2 & 55.4 & 56.8 & 30.2 & 33.9 & 34.9 \\
        CodeQwen & 46.5 & 54.9 & 56.4 & 22.5 & 26.1 & 28.0 \\
        ChipNeMo & 43.4 & - & - & 22.4 & - & - \\
        Thakur et al. & 44.0 & 52.6 & 59.2 & 30.3 & 43.9 & 49.6 \\
        VerilogEval & 46.2 & 67.3 & 73.7 & 28.8 & 45.9 & 52.3 \\
        RTLCoder-Mistral & 62.5 & 72.2 & 76.6 & 36.7 & 45.5 & 49.2 \\
        RTLCoder-DeepSeek & 61.2 & 76.5 & 81.8 & 41.6 & 50.1 & 53.4 \\
        \hline
        BetterV-CodeLlama & 64.2 & 75.4 & 79.1 & 40.9 & 50.0 & 53.3 \\
        BetterV-DeepSeek & 67.8 & 79.1 & 84.0 & 45.9 & 53.3 & 57.6 \\
        BetterV-CodeQwen & \textbf{68.1} & \textbf{79.4} & \textbf{84.5} & \textbf{46.1} & 53.7 & 58.2 \\
        \bottomrule
    \end{tabular}
    }
\end{table}

\subsection{Functional Correctness}
\label{sec:exp_2}

For functional correctness, the target is to correctly complete the Verilog module given the problem description and module definition.
The performance of BetterV is compared with GPT-3.5, GPT-4, CodeLlama-7B-Instruct~\cite{roziere2023code}, DeepSeek-Coder-6.7b-Instruct~\cite{guo2024deepseek}, CodeQwen1.5-7B-Chat~\cite{bai2023qwen}, ChipNeMo~\cite{liu2023chipnemo}, Thakur et al.~\cite{thakur2023benchmarking}, VerilogEval~\cite{liu2023verilogeval} and RTLCoder~\cite{liu2023rtlcoder}.
As shown in \Cref{tab:func_1}, the results demonstrate that our BetterV-CodeQwen have achieved the state-of-the-art performance on VerilogEval, where the pass@1 on VerilogEval-machine and VerilogEval-human outperforms GPT-4 by $8.1$ and $2.6$, respectively.
It should be noted that the pre-trained model CodeLlama-7B-Instruct has the lowest performance, but after instruct-tuning and the guidance of discriminator a huge performance is improved.

\subsection{Customized Generation in BetterV}
Besides the measurement on functional correctness, it is important to consider the performance in EDA downstream tasks, which is our customized generation. 
Note that in the rest of the paper, we conduct the experiments only based on the model CodeLlama-7B-Instruct~\cite{roziere2023code} to demonstrate our results.
Therefore, ``\textbf{BetterV}'' is employed to replace ``BetterV-Codellama'', i.e. the model fine-tuned on CodeLlama-7B-Instruct~\cite{roziere2023code}.
And ``\textbf{BetterV-base}'' refers to the base models of ``BetterV-CodeLlama'', i.e. the model has undergone instruction tuning but have not yet enhanced by discriminator.

\subsubsection{\textbf{Synthesis Nodes Reduction}}
\label{sec:exp_3}

\begin{figure}[tb!]
    \centering
    \includegraphics[width=1.0\linewidth]{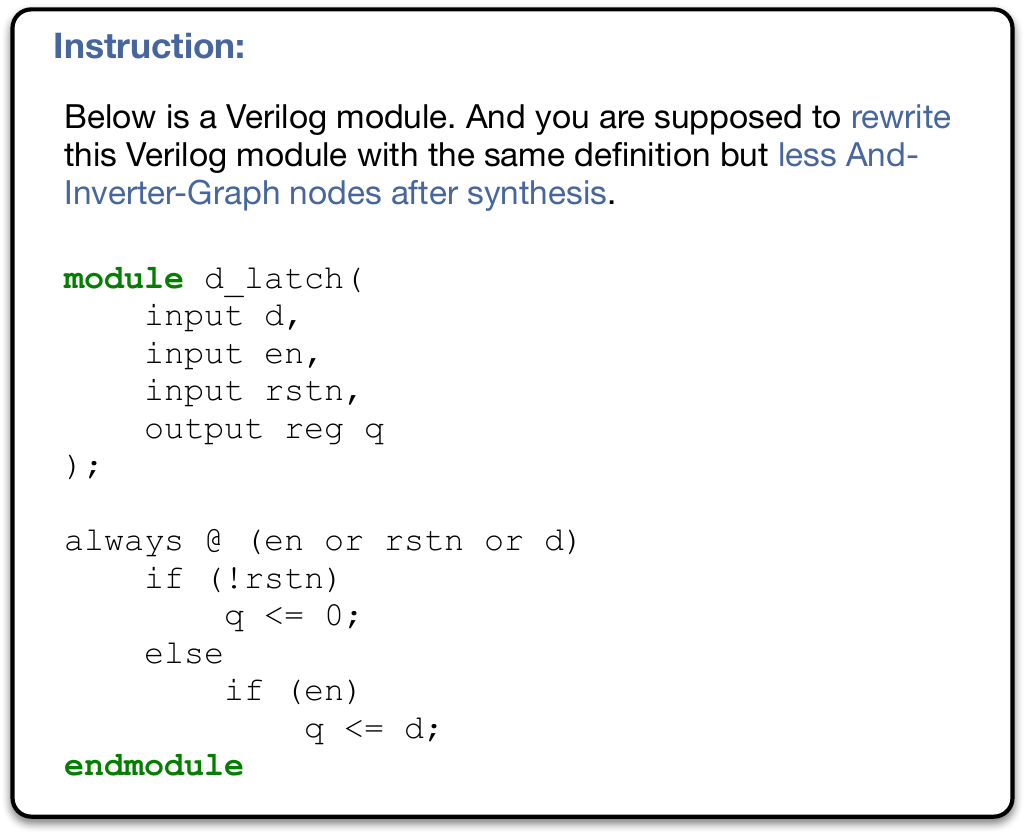} 
    \caption{
    An example to instruct the LLMs to rewrite the Verilog module to reduce the AIG nodes after synthesis.
    }
    \label{fig:rewrite}
\end{figure}

Firstly, we need to consider the hardware attributes.
Therefore, we train the discriminator to improve the performance of Verilog related to later EDA stage, i.e. the synthesis.
We make the labels for discriminator depending on whether the logic networks that synthesised from the generated Verilog have less nodes than the reference Verilog.
The netlist nodes number are collected by using the ``\code{proc; aigmap; stat}'' commands from \tool{Yosys}, which transform the Verilog into an And-Inverter-Graph (AIG).
The instruction becomes rewriting the reference to have less nodes after synthesis as shown in \Cref{fig:rewrite}.
We select several problems in VerilogEval-human to intuitively demonstrate the performance of BetterV, i.e. the reference nodes number and the generated nodes number are represented.
The presented nodes number in the table are the average nodes number of the generated Verilog.
As shown in \Cref{tab:synthesis_1}, with the guidance from discriminator, BetterV can always generate Verilog that has less netlist nodes.
The last two columns refer to the proportion of improvements in node reduction compared to BetterV-Base (``Com Base'') and Reference (``Com Ref'') respectively. It can be found that, on average, the Verilog generated by BetterV have $46.52\%$ fewer nodes than the reference model and $31.68\%$ less nodes than BetterV-base. 

The success of BetterV on the node number reduction after synthesis is significant, since it marks that LLMs start to participate in optimizing the PPA (Power, Performance, Area) of circuit design in EDA flow.
Furthermore, optimizing the Verilog implementation at the beginning of the EDA flow has proved to have meaningful influence on later stages.

\begin{table}[htb]
    \centering
    \caption{Synthesis nodes reduction with discriminator.}
    \label{tab:synthesis_1}
    \renewcommand{\arraystretch}{1.18}
    \resizebox{1.0\linewidth}{!}{
    \begin{tabular}{c|ccccc}
        \toprule
        Problem & Ref & BetterV-base & BetterV & Com Base  & Com Ref \\ \midrule
        ece241\_2013\_q8 & 657 & 333.5 & 255.3 & 23.44\% & 61.14\% \\
        m2041\_q6 & 1370 & 692.7 & 685.6 & 1.03\% & 49.95\% \\
        counter\_2bc & 673 & 666.2 & 518.9 & 22.11\% & 22.89\% \\
        review2015\_count1k & 487 & 493.4 & 402.6 & 18.44\% & 17.33\% \\
        timer & 498 & 294.3 & 247.3 & 15.97\% & 50.34\% \\
        edgedetect2 & 58 & 189.9 & 47.4 & 75.03\% & 18.27\% \\
        counter1to10 & 325 & 266.3 & 240.3 & 9.76\% & 26.06\% \\
        2013\_q2afsm & 826 & 308.8 & 296.6 & 3.95\% & 64.09\% \\
        dff8p & 50 & 42.3 & 37.8 & 10.63\% & 24.4\% \\
        fsm3comb & 844 & 167.9 & 104.4 & 37.82\% & 87.63\% \\
        rule90 & 6651 & 12435.6 & 4536.9 & 63.52\% & 31.79\% \\
        mux256to1v & 2376 & 2439.6 & 557.2 & 77.16\% & 76.54\% \\
        fsm2 & 389 & 186.53 & 121.9 & 34.65\% & 68.66\% \\
        fsm2s & 396 & 163.7 & 144.1 & 11.97\% & 63.61\% \\
        ece241\_2013\_q4 & 2222 & 1789.5 & 897.4 & 49.85\% & 59.61\% \\
        conwaylife & 43794 & 547400.3 & 27037.4 & 95.06\% & 38.26\% \\
        count\_clock & 3187 & 2497.5 & 2222.2 & 11.02\% & 30.27\%\\
        countbcd & 1589 & 932.0 & 849.3 & 8.87\% & 46.55\% \\ \bottomrule
    \end{tabular}
    }
\end{table}

\begin{table}[tb!]
    \centering
    \caption{Verification runtime reduction with discriminator.}
    \label{tab:verify_1}
    \renewcommand{\arraystretch}{1.18}
    \resizebox{1.0\linewidth}{!}{
    \begin{tabular}{c|ccccc}
        \toprule
        \multirow{2}{*}{Design} & Ref & BetterV-base & BetterV  & \multirow{2}{*}{Com Base} & \multirow{2}{*}{Com Ref} \\
        & (s) & (s) & (s) & &   \\ \midrule
         b03 & 1.233 & 1.252 & 0.857 & 31.54\% & 30.49\% \\
         b06 & 0.099 & 0.083 & 0.078 & 6.02\% & 21.21\% \\
         Spinner & 1.577 & 1.343 & 1.064 & 20.77\% & 32.53\% \\
         traffic\_light\_example & 0.583 & 0.497 & 0.480 & 3.42\% & 17.67\% \\
         Rotate & 1.153 & 1.126 & 1.034 & 8.17\% & 10.32\% \\ \bottomrule
    \end{tabular}
    }
\end{table}

\subsubsection{\textbf{Verification Runtime Reduction}}
\label{sec:exp_4}

BetterV can not only deal with the problem in synthesis, but also participate in the optimization on formal verification.
In this section, we discuss the improvement on Boolean Satisfiability (SAT) runtime reduction when doing formal verification on the Verilog by rewriting the Verilog implementation.
If we consider the concrete stages during the EDA flow, such as synthesis in last section, as the key for the circuit (PPA) performance, then the verification process decides how safe to produce a circuit.
Since the engineers always spend plenty of time to carefully verify the circuit designs, the runtime of formal verification is always a bottleneck to improve its scalability.
Therefore, it is essential to consider the impact of Verilog implementation to formal verification runtime.
Since using the SAT is one of the mainstream of formal verification, we conduct experiments to reduce the SAT solving time.

We make the labels for discriminator depending on whether generated SystemVerilog has less SAT solving time than the reference.
The SAT solving time is collected by using the ``\code{hierarchy; proc; opt; sat -verify -seq 100 -tempinduct -prove-asserts}'' commands from \tool{Yosys}, which solve the SAT problem to prove all the asserts in a circuit with 100 time steps.
The instruction becomes rewriting the reference to reduce verification runtime for solving SAT problems, which is the same as the example shown in \Cref{fig:rewrite} after replacing part of the description to "but less Boolean Satisfiability (SAT) solving time".
Since the VerilogEval benchmark doesn't contain SystemVerilog that includes the assertions inside the design, we choose another benchmark for our experiment in this section.
ANSI-C benchmarks give Verilog with safety assertions and can be used for the evaluation of BetterV~\cite{mkm2015}.
Some designs among them are selected to evaluate the performance of BetterV. Note that for the ``Rotate'' case, we uncomment all the assertions to enhance the difficulty.
As shown in \Cref{tab:verify_1}, after the discrimiative guidance, BetterV is able to further decrease the SAT solving time. We also show the improvement ratio between the verification time of Verilog generated by BetterV and BetterV-base (``Com Base'') and reference Verilog (``Com Ref'') in the last two columns respectively. It shows that the Verilog generated by BetterV can save $22.45\%$ in verification time compared with the reference Verilog, and $13.99\%$ in time compared with BetterV-base. 
The ability to reduce the verification runtime indicates that LLMs can understand how to rewrite the implementation to accelerate the provement.
And it gives the hope that we will be able to solve more complex and complicated problems and hence facilitate the issue of scalability in formal verification.

\subsection{Abltion Study}
\label{sec:exp_5}
\subsubsection{\textbf{Impact of Discriminator}}

We also show that the generative discriminator has the ability to further enhance the performance after the guidance.
In the task of functional correctness, the labels of discriminator are made by checking whether the generated modules are functional equivalence with the reference module.
The equivalence checking is done by the \tool{eqy} tool in \tool{Yosys}.
With the problems in VerilogEval-human, the results in \Cref{tab:func_2} illustrate that the guidance from the discriminator can not only enhance the capability on our fine-tuned LLMs, i.e. these base models, but also on the original pre-trained LLMs, i.e. CodeLlama.
This observation indicates that our trained discriminator can be employed to any LLM, only if they have the same vocabulary size.

For syntactic correctness, we only consider whether the syntax of the generated Verilog is corrected or not.
We construct the labels of generative discriminator based on whether the generated module can be complied by the \tool{Yosys} with the command ``\code{prep}'', which is a generic synthesis script.
We also evaluate the performance with the problems in VerilogEval-human, which is demonstrated in \Cref{tab:syntax_1} and the results show that in the task of generate syntactically correct Verilog, BetterV can also achieve remarkable performance, i.e. over $99$ pass@10.
It can also be observed that the discriminator can help largely enhance the syntactic correctness, i.e. 7.6 and 4.5 improvement on pass@1 for the original CodeLlama-7B-instruct model and the BetterV-base, respectively.

\begin{table}[tb]
    \centering
    \caption{Impact of discriminator on functional correctness.}
    \label{tab:func_2}
    \renewcommand{\arraystretch}{1.18}
    \resizebox{0.78\linewidth}{!}{
    \begin{tabular}{c|ccc}
        \toprule
        Model & pass@1 & pass@5 & pass@10 \\
         \midrule
        CodeLlama & 18.2 & 22.7 & 24.3 \\
        CodeLlama + Dis & 20.3 & 24.1 & 24.7  \\
        BetterV-CodeLlama-base & 40.0 & 49.5 & 53.0 \\
        BetterV-CodeLlama & 40.9 & 50.0 & 53.3 \\
        \bottomrule
    \end{tabular}
    }
\end{table}

\begin{table}[tb]
    \centering
    \caption{Impact of discriminator on syntactic correctness.}
    \label{tab:syntax_1}
    \renewcommand{\arraystretch}{1.18}
    \resizebox{0.78\linewidth}{!}{
    \begin{tabular}{c|ccc}
        \toprule
        Model & pass@1 & pass@5 & pass@10 \\ \midrule
        CodeLlama & 41.5 & 50.8 & 53.8 \\
        CodeLlama + Dis & 49.1 & 58.4 & 61.1  \\
        BetterV-CodeLlama-base & 82.6 & 97.6 & 99.2 \\
        BetterV-CodeLlama & 87.1 & 98.2 & 99.3 \\ 
        \bottomrule
    \end{tabular}
    }
\end{table}
\section{Discussion}

\subsection{Current limitations}

We have shown that BetterV can achieve remarkable performance on Verilog generation and the generative discriminator enables improvements on Verilog-related EDA downstream tasks.
While BetterV contributes valuable insights and a practical method for enhancing the performance of open-source LLMs, it still has limitations on closed-source models due to the need for access to token-level probabilities. 
The research about guidance generation on closed-source models remains an unsolved issue.

In addition, while the generative discriminator uses only a small model, the additional computational overhead it brings cannot be ignored.
Research to reduce this overhead is needed to accelerate the guidance in an accuracy-optimal manner.
For example, it is possible to use model inference acceleration methods such as quantization and pruning on the discriminator.

\subsection{Applicability to other domains}

The potential applicability of our methods to other domains with similar challenges of data scarcity and precision requirements is promising.
For example, many domains utilize structured languages (e.g., SQL for databases, LaTeX for typesetting, and XML for web services). 
The principles behind BetterV's instruct-tuning and generative discriminators can be adapted to improve the generation of such languages, particularly where precision and correctness are non-negotiable.
Fields such as genomics, where data is highly structured and errors can have significant consequences, could benefit from our approach. 
The instruct-tuning methodology could be adapted to understand the syntax and semantics of genetic sequences, while generative discriminators could be used to guide the generation of sequences to complex biological constraints.

\section{Conclusion}

In summary, this paper introduces a novel framework, BetterV, for Verilog generation, to control the Verilog implementation and optimize its performance on various aspects.
We introduce a complete and easy-following way to collect and process Verilog data.
Our domain-specific instruct-tuning successfully teaches the large language models (LLMs) the knowledge of Verilog with the help of our processed data from the internet.
A data augmentation process is proposed to further enhance the diversity of dataset.
Particular generative discriminators are then trained to meet specific requirements for different downstream tasks in the electronic design automation (EDA) flow.
The experimental results show the state-of-the-art capability of BetterV on various tasks.
BetterV marks a pioneering advancement on optimizing the PPA (Power, Performance, Area) of circuit design and introduces a promising direction to accelerate the verification process.
Future work could explore additional domain-specific adaptations and incorporate other powerful techniques in LLMs to further enhance BetterV's performance and applicability.


\newpage





\end{document}